# A Multimodal, Multitask System for Generating E-Commerce Text Listings from Images

Nayan Kumar Singh

Independent Researcher, Bangalore, India Email: nayan.ksingh.r@gmail.com

## Abstract

Manually generating catchy descriptions and names is labour intensive and a slow process for retailers. Although generative AI provides an automation solution in form of Vision-to-Language Models (VLM), the current VLMs are prone to factual "hallucinations". Siloed, single-task models are not only inefficient but also fail to capture interdependent relationships between features. To address these challenges, we propose an end-to-end, multi-task system that generates factually-grounded textual listings from a single image. The contributions of this study are two proposals for the model architecture. First, application of multi-task learning approach for fine-tuning a vision encoder where a single vision backbone is jointly trained on attribute prediction such as color, hemline and neck style and price regression. Second, introduction of a hierarchical generation process where the model's own predicted attributes are embedded in a prompt and fed to the text decoder to improve factual consistency. The experiments demonstrate the superiority of this architecture. The multi-tasking approach outperforms both the independent price regression, with a 3.6% better $R^2$ Value and attribute classification, with a 6.6% improvement F1-score. Critically, the hierarchical generation process proves highly effective, slashing the factual hallucination rate from 12.7% to 7.1%—a 44.5% relative reduction—compared to a non-hierarchical ablation. The hierarchical approach also reduces the latency of the autoregressive text generation process by a factor of 3.5 when compared to direct vision-to-language model of similar size. One minor caveat is that the model does perform 3.5% worse than direct vision-to-language model on ROUGE-L score.

# 1. Introduction

Data analysis of current portfolio and competitors offering in highly dynamic e-commerce clothing segment is task of fundamental importance for clothing designers. Along with this listing the product with catchy name and description is of paramount importance for better reach and sales. Manually naming and describing each product is both time consuming and difficult to scale.

Generative AI is emerging as a potential solution for this problem. However, there are several architectures being employed for image captioning. Some systems treat the constituent tasks of listing creation as an independent task. Thus, all tasks such as attribute tagging, description writing, and price estimation have their siloed workflows. These models don't capture the holistic product understanding. Others, use Vision-Language Models (VLMs) for the task. Although these models are capable of generating fluent text, they are prone to factual "hallucinations". They can invent some details that are not present in the image. (Jiang et al., 2024) This requires expensive manual scrutiny and correct cycles.

To address these challenges of architectural fragmentation and factual inconsistency, we propose holistic multi-task learning based hierarchical model. This system is built upon two core, synergistic contributions:

- The Hierarchical Generation (HG) process is introduced to reduce factual hallucinations. The model first predicts a set of structured attributes and then uses these attributes to constrain a large language model by a prompt. This ensures that the generated text remains faithful to the visual evidence.
- To get a holistic understanding of the product and capture subtle visual cues that determine product value we employ multi task learning approach (Ruder, 2017). A single vision encoder is jointly optimized on attribute prediction and price estimation, forcing it to learn a rich, shared representation that benefits all downstream tasks.

The primary contributions of this paper are listed below:

1. A hierarchical generation process that uses predicted structured data to guide text generation, significantly improving the factual consistency of product descriptions.

2. A holistic visual representation learned via multi-task learning that jointly models product attributes and price, leading to improved predictive accuracy and model efficiency.

3. Comprehensive analysis of existing methodologies for clothing image to text generation

# 2. Related Work

Our research builds upon established work in multimodal learning, controlled text generation, and multi-task learning.

## 2.1 Multimodal Learning for E-commerce

The application of machine learning in e-commerce has evolved from unimodal to multimodal approaches. Early work often focused on single-modality tasks, such as product classification from text or visual-based recommendation systems (He et al., 2016). Recognizing these limitations, more recent work has focused on the synergy between modalities, as multimodal frameworks demonstrate higher predictive accuracy (Gao et al., 2023). This has led to three key research thrusts relevant to proposal: visual attribute extraction, which models the problem as a multi-label classification task (Chen et al., 2022); image-to-text generation, which leverages VLMs to produce descriptive captions (Vinyals et al., 2015); and price prediction from visual cues, where models learn to estimate market value by capturing subtle indicators of quality and style (Bell & Pádraic, 2016). To my knowledge, the proposed system is the first system to unify these three tasks into a single, end-to-end architecture designed to improve factual consistency through its multi-task, hierarchical approach.

## 2.2 Hierarchical and Controlled Text Generation

A primary challenge for LLMs is maintaining factual grounding. To mitigate hallucinations, several control strategies have been developed. Retrieval-Augmented Generation (RAG) grounds model outputs by first retrieving relevant documents from an external knowledge base to provide context (Lewis et al., 2020). Chain-of-thought prompting improves logical consistency by guiding the model to generate intermediate reasoning steps (Wei et al., 2022). The proposed Hierarchical Generation (HG) method aligns with this paradigm of controlled generation. By first predicting structured attributes and then using them to condition the language model, we use the image's content as a self-contained, verifiable knowledge source to guide the final text output, directly addressing the hallucination problem in a novel way.

## 2.3 Multi-Task Learning in Computer Vision

Multi-Task Learning (MTL) is a training paradigm where a model learns multiple objectives from a shared representation, often leading to improved data efficiency and generalization by encouraging the model to learn features that are broadly useful (Ruder, 2017). The assumption is that an inductive bias introduced by the auxiliary tasks can help the main task. MTL has been successfully applied across numerous domains, including autonomous driving for joint object detection and segmentation (Teichmann et al., 2018). We apply this principle to learn a Holistic Visual Representation (HVR), positing that the tasks of predicting visual attributes and estimating price are sufficiently related—both relying on latent features of quality, style, and material—to produce a more powerful shared representation of a product's visual identity and market value.

# 3. Methodology

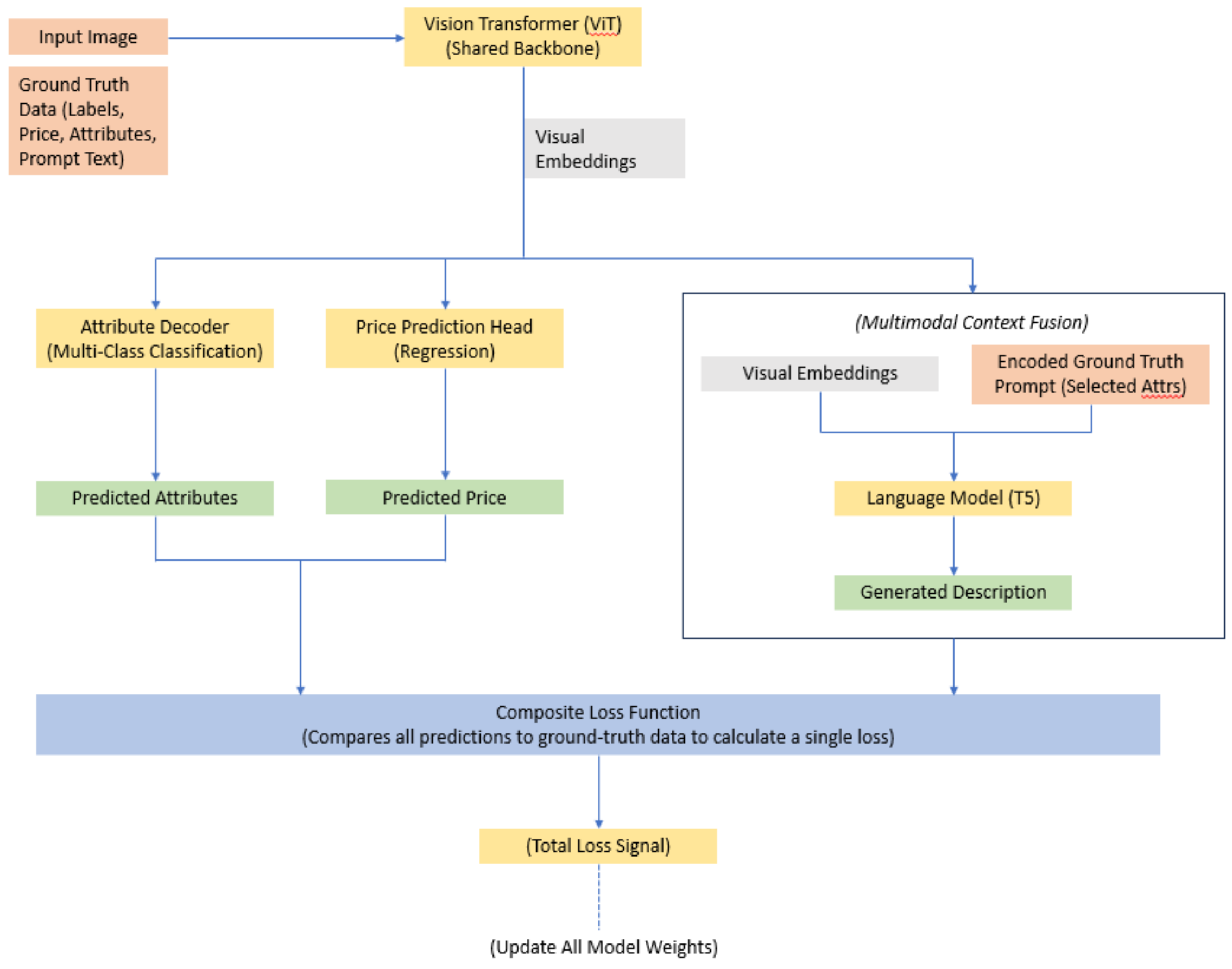

*Figure 1: Multi-Task learning – Hierarchical Model architecture during training*

## 3.1 The Encoder

The proposed model architecture, as shown in Figure 1, relies on a vision encoder and a text decoder structure. The encoder chosen for the study is Vision Transformer, hereafter ViT (Dosovitskiy et al., 2020). ViT relies on self-attention mechanism that is , in theory, better than traditional CNNs to capture fine grained local features such as fabric texture or holistic context defining the product style such as overall silhouette.

Among the various attributes, a good mix of visually detectable attributes such as colour, sleeve length etc. and derived attributes such as style, occasion etc. was chosen. This deliberate selection of attributes is intended to enable the model to learn not only "what's visible" but also "what it implies". For example, if model frequently encounters black one-piece (visually identifiable cues) it can attribute it to evening dress suitable for parties, as shown in Figure 2.

The ViT model processes each image as a hidden state for each image patch, and also provides a pooled representation for the entire image. We take this pooled representation as the input for the task specific heads. As shown in the Figure 1, two different MLP heads is attached onto the visual embeddings to process the attributes and price respectively.

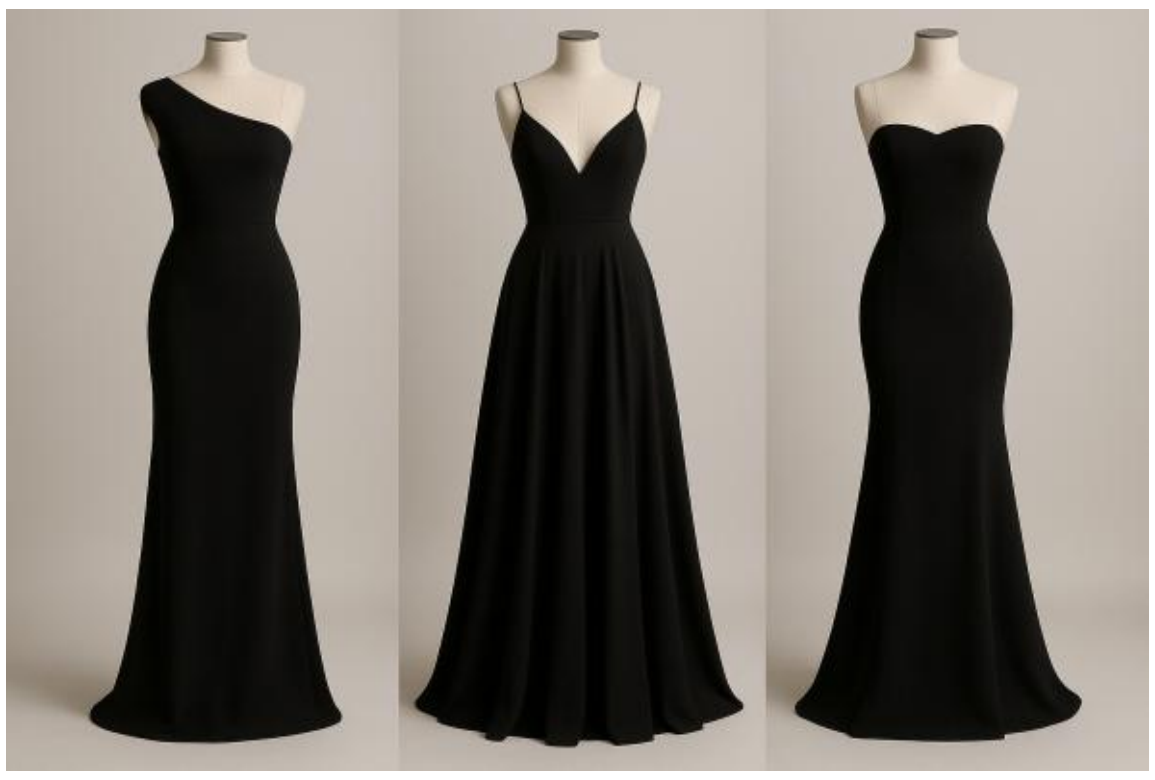

*Figure 2: Three types of formal/ evening dresses, model captures usual pattern like sleeveless and floor length dress for mapping*

*Table 1: Attributes detail available for dresses shown in Figure 2*

| **Attribute** | **Dress 1** | **Dress 2** | **Dress 3** |
|---|---|---|---|
| Neck | One Shoulder | Deep V-Neck | Sweetheart Neck |
| Sleeve Length | **Sleeveless** | **Sleeveless (Spaghetti Straps)** | **Strapless** |
| Print or Pattern Type | Solid | Solid | Solid |
| Type | Bodycon Gown | A-Line Gown | Mermaid Gown |
| Hemline | Flared | Flared | Flared |
| Pattern | Plain | Plain | Plain |
| Length | **Floor Length** | **Floor Length** | **Floor Length** |
| Sleeve Styling | No Sleeves | Strappy Sleeves | Strapless |
| Ornamentation | None | None | None |
| Occasion | **Evening / Formal** | **Evening / Formal** | **Evening / Formal** |
| Fabric | Stretch Crepe | Satin or Crepe | Stretch Crepe |
| Fit | Slim Fit | Regular Fit | Body Fit |
| Colour | Black | Black | Black |

## 3.2 The Decoder

Inputs -

1. The final hidden state of ViT is extracted. This vector is projected to suitable dimension for decoder input, 768 -> 512. We have used T5 small as text decoder.

2. A generic text prompt created from a few ground truth attributes is also passed to the decoder. These are created as text tokens e.g., *"generate a product listing for an item of type 'garment', suitable for 'everyday wear'"*.

```
occasion = row.get('Occasion', 'everyday wear')
item_type = row.get('Type', 'garment')
prompt = f"generate a product listing for an item of type '{item_type}', suitable for '{occasion}'"
```

Both the visual token and the encoded prompt tokens are concatenated along the sequence dimension. This creates a single, fused context vector (e.g., [IMAGE_TOKEN, PROMPT_TOKEN_1, ...]) that the T5 decoder can attend to. The idea to let the model learn integration of high-level textual instruction with low level visual details. We intentionally didn't provide detailed visual attributes otherwise model will become lazy and only rephrase the attributes and ignore visual embedding.

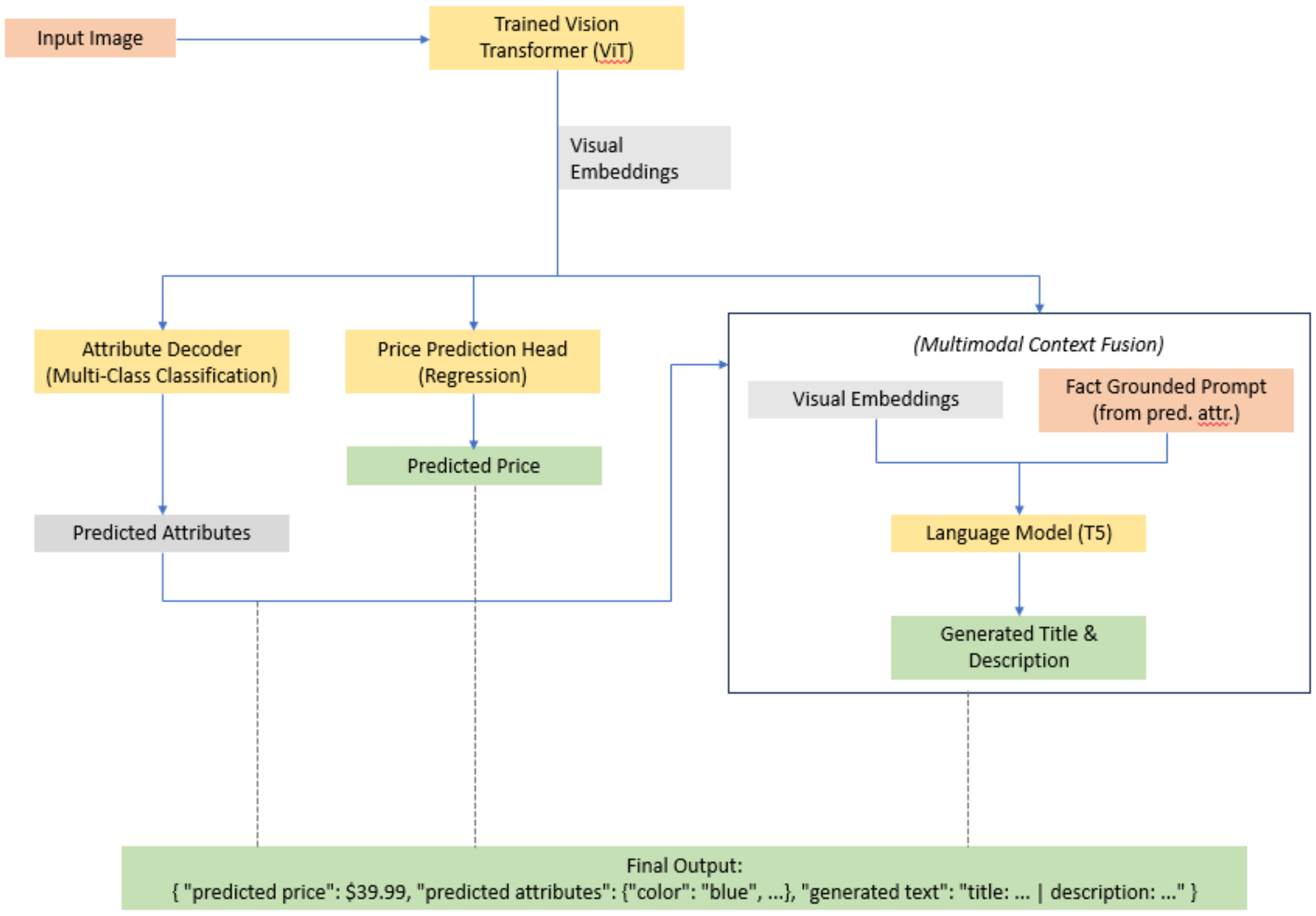

*Figure 3: Multi-Task learning – Hierarchical Model architecture during inference*

A different strategy is employed during inference, as shown in Figure 3. The classification head is asked for product attributes. This predicted attribute is transferred to the text generator using a detailed prompt. The T5 decoder uses the visual embedding and the new factual grounded prompt to return the final description and name of the listing. The proposed grounding should help with lowering the hallucination.

## 3.3 Joint Loss Function

The model is jointly optimized for the three tasks together. During training the losses from each task is combined together to form the total loss. The loss from classification (Cross-Entropy) and from regression (MSE) can vary hugely in magnitudes. To prevent one head dominance over the other, they are weighted before summation.

Mathematically,

$$L_{\text{total}} = \alpha \cdot L_{\text{attr}} + \beta \cdot L_{\text{price}} + \gamma \cdot L_{\text{text}}$$

Individual Losses:

1. Price Loss: Mean Squared Error (MSE) loss calculated between the Predicted Price and the ground-truth price.
2. Attribute Loss: Cross Entropy Loss calculated for each of the 12 selected attributes between the predicted logits and the ground-truth labels. The final attribute loss is the average of these individual losses.
3. Text Loss: The standard cross-entropy loss for language modelling is computed internally by the T5 model when labels are provided.

The gradient from total loss backpropagates through all model components, forcing the ViT encoder to learn a feature representation that is beneficial for all three tasks simultaneously. This multitasks learning approach, in theory, incentivizes the model to learn the relationship between attributes and market value. This is demonstrated in Figure 4.

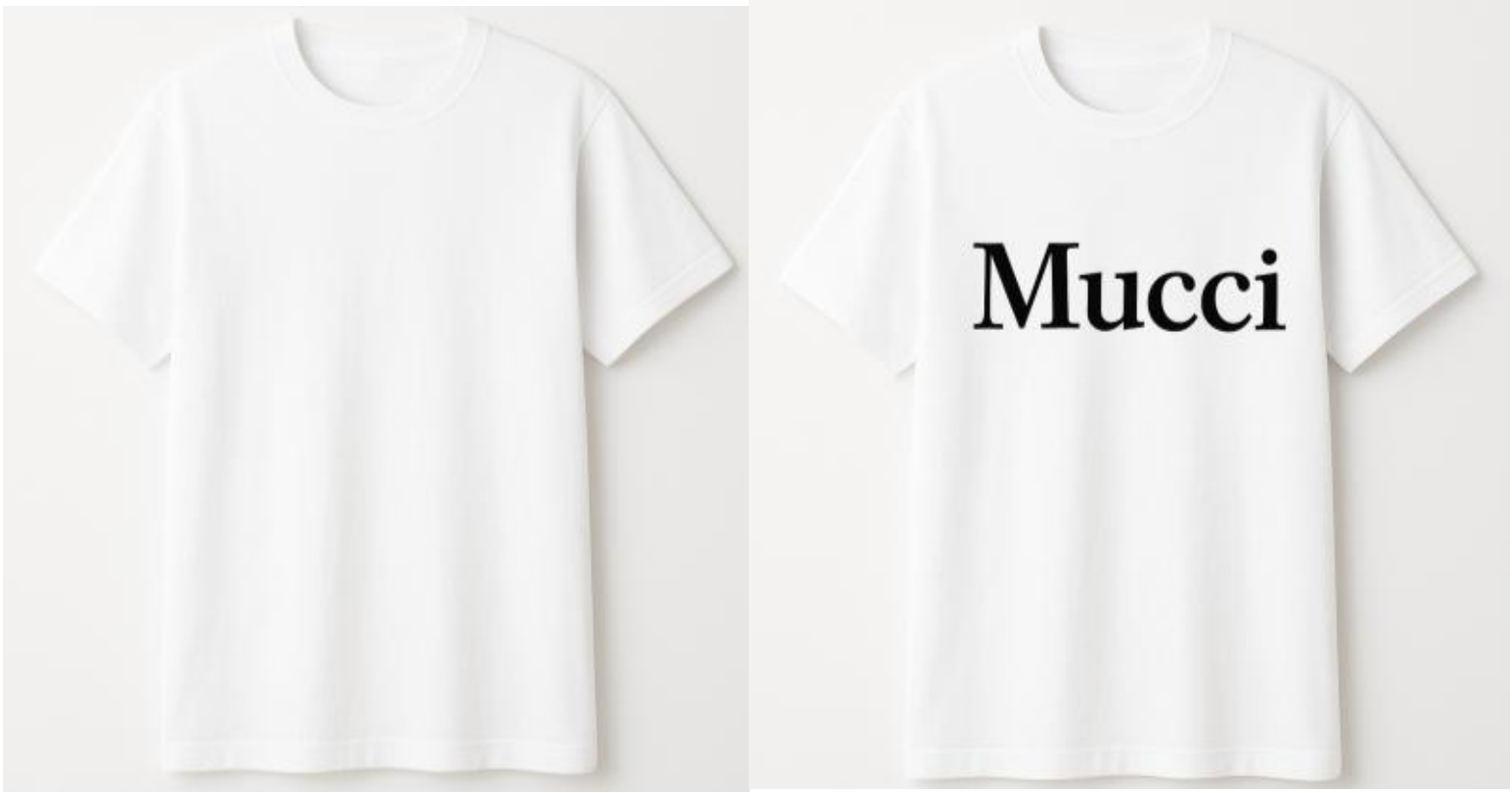

*Figure 4: Which one looks expensive? The plain white shirt with exact same attributes may be much cheaper than a shirt printed with some characters. The training wants to simulate learning this relationship.*

# 4. Experiments

## 4.1 Dataset

The dataset consists of 14.2k women clothing listings collected from several public e commerce website. The data is then tabulated using their features. This includes a set of 52 different columns including rating, price, name and description. The images were resized to 224 x 224 pixels. These 52 inputs were segregated in three different types of features – easy visually detectable like neck shape, could be detected like fabric texture and inferred features like occasion. This mix was considered for training the model. Other features that were too niche for model to learn like brand or unique identifiers were removed. Final attributes selected for training is listed below:

```
SELECTED_ATTRIBUTES = [
  # --- Tier 1: Core Visuals ---
  'Neck',
  'Sleeve Length',
  'Print or Pattern Type',
  'Type',
  'Hemline',
  'Pattern',
  # --- Tier 2: Detailed Visuals ---
  'Length',
  'Sleeve Styling',
  'Ornamentation',
  # --- Tier 3: Contextual for Text ---
  'Occasion',
  'Fabric',
  'Fit'
```

## 4.2 Evaluation Metrics

The following metrics were employed to determine the model's performance.

**Attribute Prediction:** The multi-classification task is evaluated using Macro F1 Score. This score was used because the it provides a balanced assessment across all attribute classes, regardless of their frequency.

**Price Prediction:** For this regression task, we report Mean Absolute Error (MAE) for its intuitive interpretation in currency units, Root Mean Squared Error (RMSE) to penalize large prediction errors, and R-squared ($R^2$) to measure the proportion of the variance in the price that is predictable from the model's inputs.

**Text Generation:** We use a suite of metrics to assess text quality:

- BLEU-4 and ROUGE (1, 2, L): We report these standard n-gram overlap metrics for comparability with prior work in text generation.
- Hallucination Rate: This is the primary metric for factual consistency, directly evaluating the HG claim. It is defined as the percentage of generated descriptions where the model mentions an attribute value that contradicts the structured attributes predicted by its own vision module. This measures the model's internal consistency. A lower rate is better.
- Model Efficiency: To evaluate the computational cost of the models, we report:

- Parameters (M): The total number of trainable parameters in millions, indicating the model's size.
- Latency (ms): The average time in milliseconds required to perform a full end-to-end inference for a single sample on the target hardware.

## 4.3 Baselines and Ablations

To prove that proposed multi task learning hierarchical architecture is better suited for this task, the model is compared against several well-defined baselines.

**Baseline-Siloed:** These category of model consists of two models independently trained for price prediction and attribute prediction task respectively. These models are required for testing the impact of multi- task learning approach. If the proposed solution provides better performance, then these independent frameworks then the proposed model was able to learn the relationship between the attributes and price.

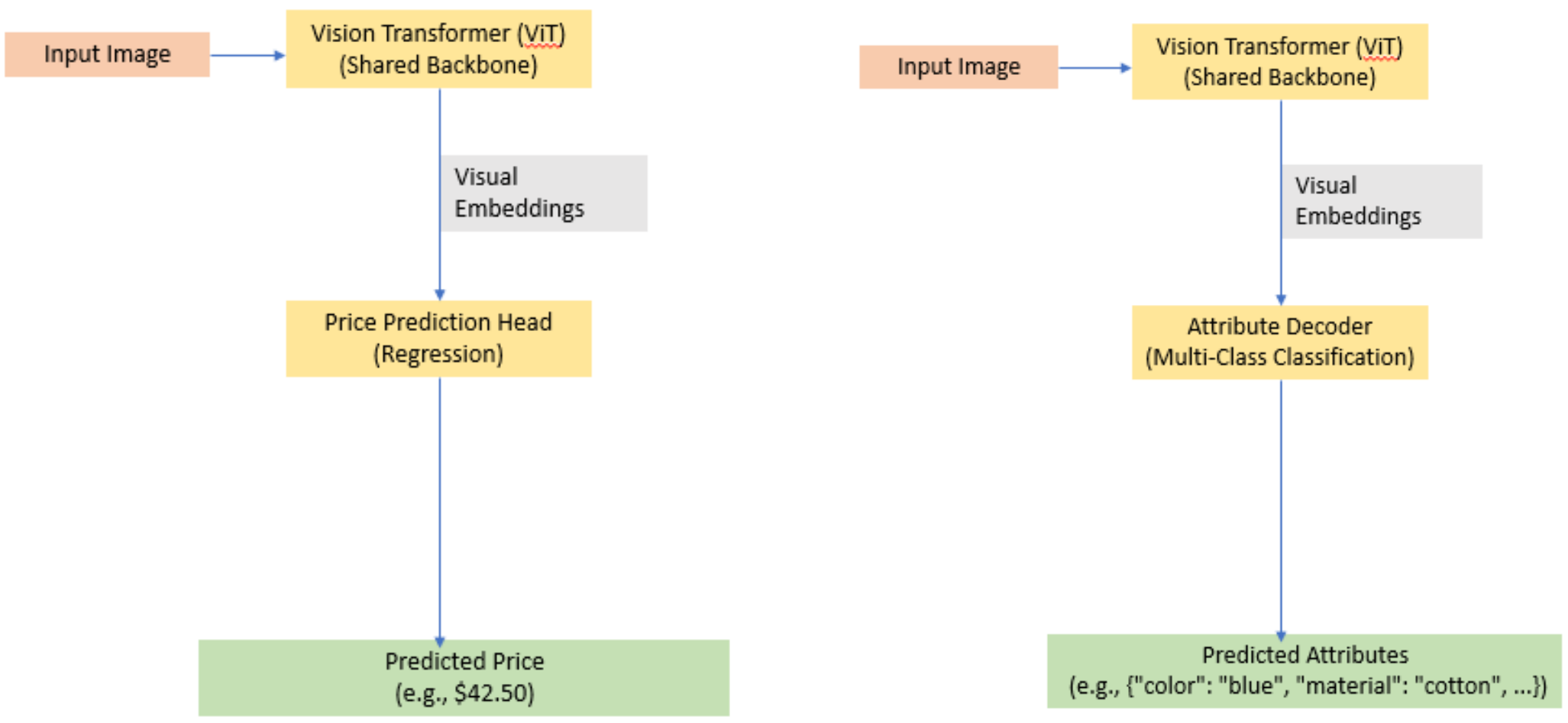

*Figure 5: Model Architecture of Siloed Price Model and Siloed Attributes Model*

**Baseline-Direct VLM:** The direct Vision-to-Language model takes in input image, generate the entire product description directly from the image, without any intermediate structured attribute prediction. This is the primary baseline for evaluating the hierarchical generation claim. Two different VLM model types were chosen as baseline. A single unified transformer structure the GIT-base model fine tuned on the training dataset. The decoder attends to image patches and the previous text. A classic example of Self-Attention mechanism. The other type with ViT encoder and T5 small decoder. Here, the decoder attends to only the output of encoder, an example of cross-attention mechanism. The pre-trained model taken as reference needs to be of similar parameter size as the baseline proposed model. This is necessary requirement as we don't the reference model to simply brute force the architectural limitation with model size or pre-training data. Consequently, the GIT-base model with ~138M parameters was selected.

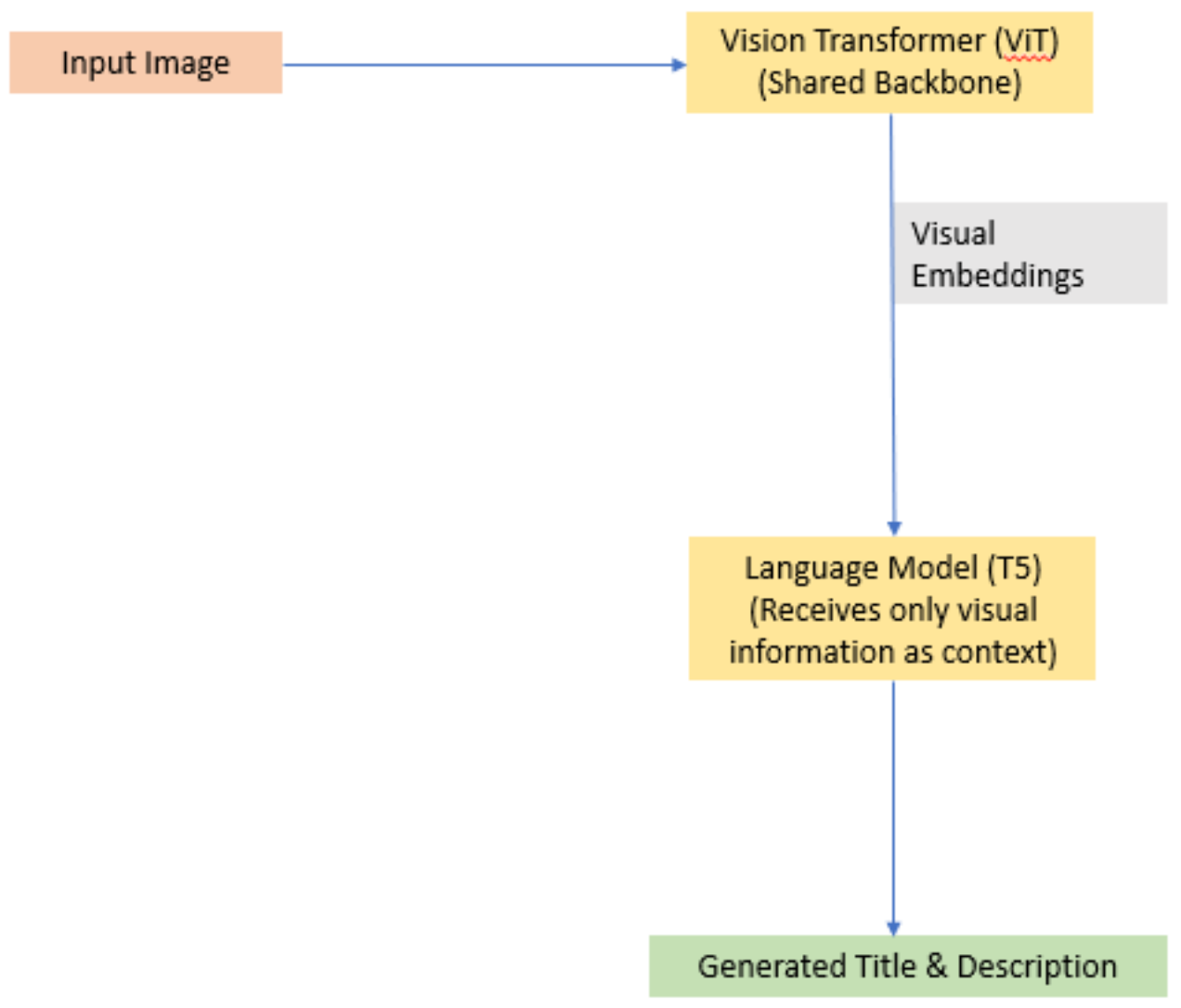

*Figure 6: Model Architecture of direct Vision to Language Model (VLM)*

**Baseline-Hybrid:** This model has two different models trained independently. One model is XG Boost trained on ground truth attributes tasked to generate the price prediction. The other one the siloed attribute training model that generates attributes from image. During inference the predicted attributes by this model is fed to the XG boost model to predict the price. This model is taken as baseline to prove that information embedded in the visual embedding is necessary. The long data flow chain can omit some relevant information.

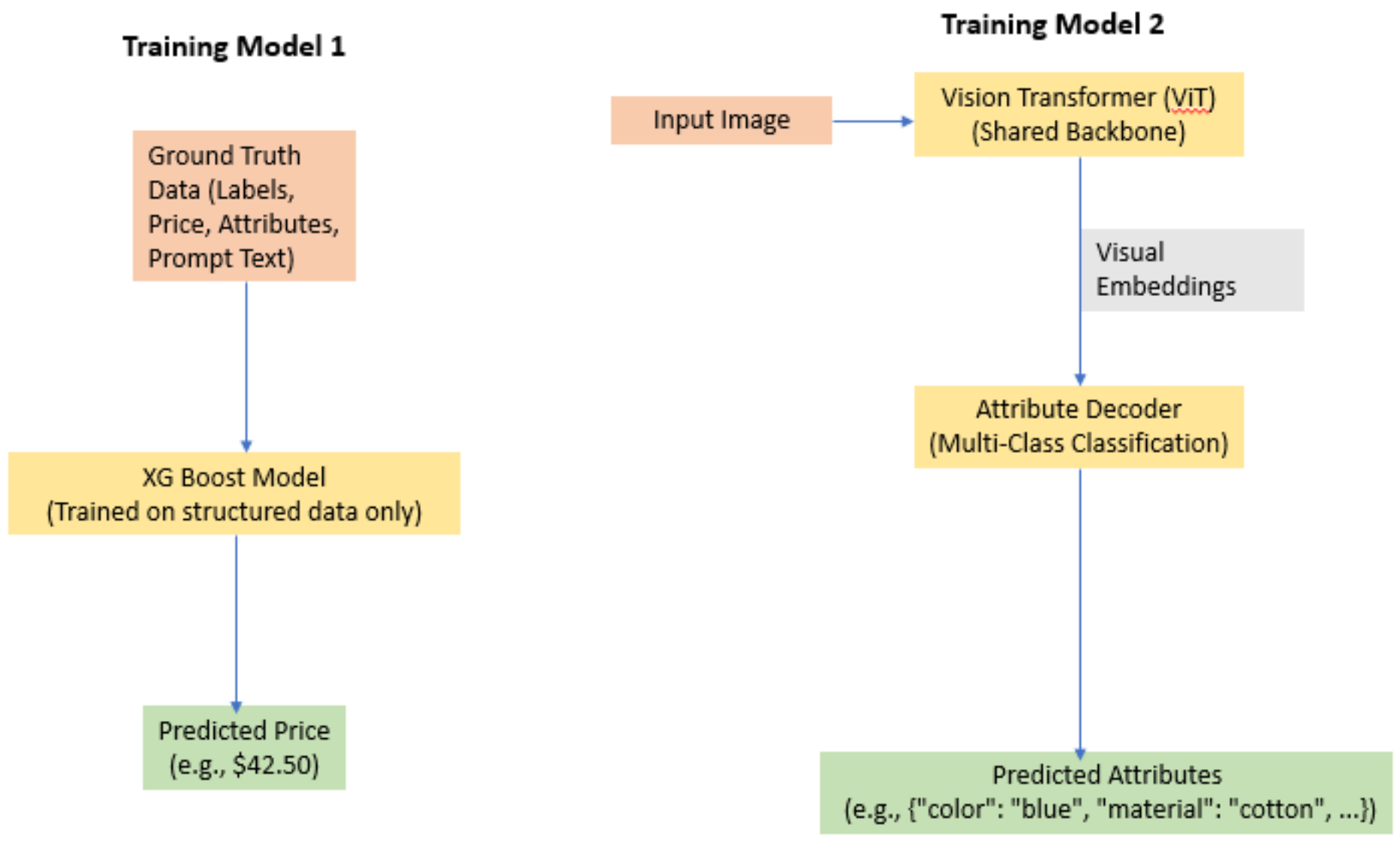

*Figure 7: Training Architecture for Hybrid Model, both models are trained independently*

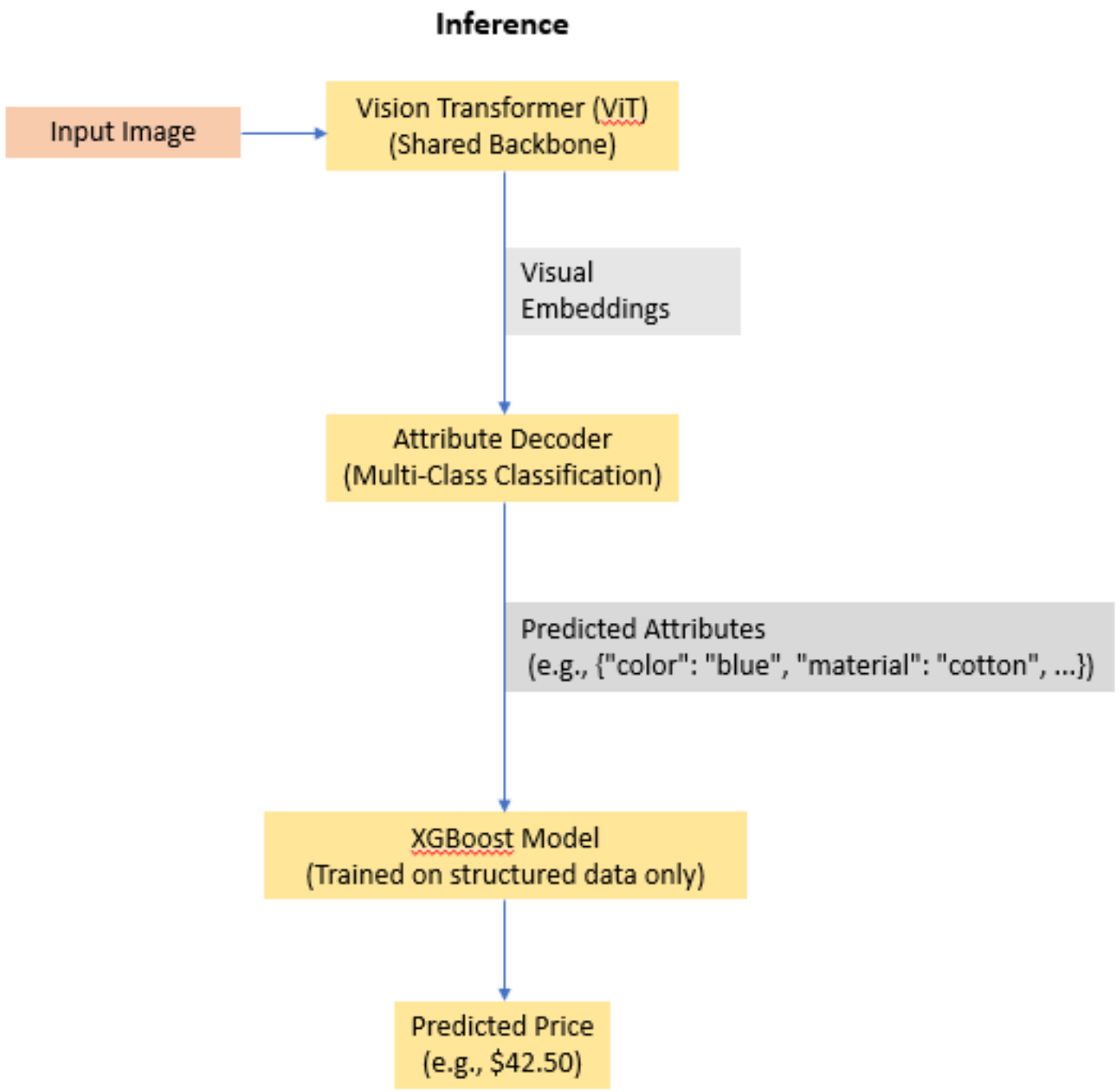

*Figure 8: Inference Architecture for Hybrid Model, independently trained model is joined together*

**Ablate-No MTL:** The same as proposed model architecture but without the price prediction head. Since, the price head is removed, the related loss is also removed from the combined loss. This model tests if learning price and attributes together aids in model performance.

**Ablate-No- Hierarchy:** The same as proposed model but during inference, the text decoder is only passed the visual embedding. The predicted attributes are not passed to the model. This model is required to test whether grounding the text decoder with predicted attributes helps in reducing hallucination or not.

# 5. Implementation Details

The proposed model consists of a ViT-Base/16 vision backbone and a T5-Small text generator. The total parameters are 147.7M. The model was trained end-to-end for 10-15 epochs using the Adam optimizer with a learning rate of 1e-4 and a batch size of 12-16. The composite loss function was weighted with $\alpha$=0.4 (attributes), $\beta$=0.1 (price), and $\gamma$=0.5 (text), values determined via an ablation study. All experiments were conducted on a single NVIDIA GeForce RTX 4050 with 6 GB of VRAM. Further details on the learning rate schedule, optimizer parameters, and beam search configuration are provided in Appendix A.

# 6. Results and Discussion

## 6.1 Multi Task Learning vs Independent Task Learning

While the results shown in Tabe 2 look abysmal for any model to be deployment ready. The task of predicting price is inherently difficult with only image as the input. The brand, often not visually detectable, plays an important role factor here. A plain white T shirt from Gucci may be 100x times more costly than H&M t shirt. The price is also influenced by factors like market trend, seller reputation etc. The model has no access to this information. Also, due to huge variation between prices – 1500INR to 25000INR, the RSME of 1935 INR, though high, is not unexpected. Even with all these influencers the proposed model is able to explain about 50% of variation in price with only images as input.

I do concede that these results are not deployment ready and we need to add more features in the model to correctly simulate the phenomena. However, the results perfectly demonstrate that multitasking is able to explain 3.64% better variance in price than the independent price model.

*Table 2: Comparison between Multi-Task Learning approach and Siloed approach for regression*

| Model | Price MAE (in INR) | Price RMSE (in INR) | Price $R^2$ |
|---|---|---|---|
| Multi Task Learning | **1067.67** | **1935.81** | **0.476** |
| Siloed - Price | 1074.31 | 1966.55 | 0.460 |

Similarly, the average macro F1 score of 0.3 is modest. It is a massive multi-label, multi-class classification problem with several inherent challenges. We are predicting the attributes for 12 categories. But each of these attributes have many different classes. Thus, each image has 399 (sum of each class under each attribute) possible labels. Also, the dataset contains a long tail of rare attribute values. Some of these labels is difficult for even humans to detect with just eyes for eg. difference between “Fabric” features labels “Cotton”, “Cotton Silk”, "Polycotton", "Organic Cotton", "Pure Cotton", "Cotton Blend" is hard to judge.

Though the absolute value of F1 score is not suitable for deployment, the trend in values lead to an import conclusion. The multi task learning model demonstrated a 6.6% better performance than the Siloed model.

*Table 3: Comparison between Multi-Task Learning approach and Siloed approach for multi-class classification*

| **Model** | **Attribute F1** |
|---|---|
| Multi Task Learning | **0.338** |
| Siloed - Attributes | 0.317 |

## 6.2 Hybrid Model vs Multi-Task Learning

As shown in Table 4, the proposed model achieved a 5.8% lower Mean Absolute and a 10.1% lower Root Mean Squared Error compared to the Hybrid Model. Also, the proposed solution explains the variance in price better by 34.9% than the hybrid model. This result can be because of data loss during the pipeline. The predictor model is not able to “see” the image. It only sees the discrete, simplified representation of that image as described by the 12 attributes. Thus, any data that is not represented under these categories is essentially lost such as craftmanship, subtle brand cues stylist nuances. For e.g. whether the garment's silhouette is elegant and well-constructed or ill-fitting influences the listing price of product.

*Table 4: Comparison between Multi-Task Learning approach and Hybrid approach for regression*

| Model | Price MAE (in INR) | Price RMSE (in INR) | Price $R^2$ |
|---|---|---|---|
| Multi Task Learning | 1067.67 | 1935.81 | 0.476 |
| Hybrid Model | 1133.66 | 2153.15 | 0.352 |

## 6.3 Model Efficiency

The proposed model has 15% fewer parameters than the both the siloed models combined. However, this can provide relevant basis for comparison because the multitask model contains parameters for text generation while the siloed models only generate the attributes or the price. Thus, the time taken is 19ms or 13ms by the siloed models or hybrid model when compared to 251ms taken by MTL model.

As seen in Table 5, the parameter count, due to the additional MLP heads for price and attribute prediction, increases marginally (~0.4M). This may add to a negligible computation cost. However, the information provided by them greatly accelerates the autoregressive text generation process. They provide the language model a concise and fact grounded prompt, thereby, significantly reducing the search space for the decoder. Thus, the beam search algorithm converges quickly, reducing the end-to-end latency by a factor of 3.5x compared to an equivalent VLM model. The GIT-base model despite its higher parameter count takes lesser time to generate the output. The unified transformer mechanism or the length of text generated may be the factor at play here. However, we will not investigate this further in this paper, as our focus remains at Multi Task Learning-Hierarchical model. The proposed model clearly surpasses its counterparts.

*Table 5: Comparison between different approaches for model size and latency*

| Model | Parameters (M) | Latency (ms) |
|---|---|---|
| Multi Task Learning - Hierarchical | 147.7 | 251 |
| ViT-T5small-Direct Vision to Language | 147.3 | 883 |
| GIT-base-Direct Vision to Language | 176.6 | 586 |
| Baseline-Siloed | 173.1 | 19 |
| Baseline-Hybrid | 86.7 | 13 |

## 6.4 Hierarchical vs Non-Hierarchical

The data presented in Table 6, is evaluated on 1422 validation set samples. The clear distinction between the performance of hierarchical model and non-hierarchical model lies in the hallucination rate metric. The results show that the proposed approach demonstrates a 44.5% relative reduction in factual contradictions. This impact can be visualised by a qualitative comparison presented in Figure 9. Thus, a containing prompt with attributes generated from visual analysis is a highly effective strategy to ensure that the final text remains grounded in visual facts. However, this improvement in factual consistency comes at the cost of n-gram based similarity metrics. Since, the non-hierarchical approach is less constrained, it performs 6.4% better on the ROUGE-L metric.

But, this trade off of similarity score for factual consistency is desired. Due to lower constraints the non-hierarchical model may be more creative. Thereby, the model can "invent" some facts to match the reference text n-grams.

The average generated length for hierarchical and non- hierarchical model is 34 words and 32 words respectively. Thus, the additional attribute prompting encourages thoroughness because we are supplying information to include in the final output.

We are not testing the hallucinate rate for direct vision to language models because these models are not tasked to predict the attributes. This is intentional decision made to ensure that VLM models are not overwhelmed with the tasks. This decision remains consistent with the objective of model to generate the name and description of the product.

The hierarchical model performs poorer than the direct VLM model in fluency metrics, e.g. 3.7% lower ROUGE-L score than the ViT-t5small VLM baseline. However, as discussed earlier this is deliberate trade off better factual consistency. If we remove the hierarchy (attribute grounding) from the model during inference the model (Multi Task Learning – Non-Hierarchical) performs 2.3% better than the ViT-t5small VLM baseline on ROUGE L score. This proves that hierarchical training is still the better approach the direct VLM architectures.

The GIT-base VLM model performs poorly on all text-based metrics. This may be because the model lacks pre-trained decoder that was specifically trained on large corpus of text like T5small. This suggests that mixing pre-trained capabilities ensures better results than relying on "purity" of training done on a smaller dataset.

*Table 6: Comparison between different approaches for factual consistency and phrase fluency*

| Model | BLEU | ROUGE-1 | ROUGE-2 | ROUGE-L | Hallucination Rate |
|---|---|---|---|---|---|
| Multi Task Learning - Hierarchical | 0.155 | 0.420 | 0.240 | 0.403 | 7.0% |
| Multi Task Learning – Non-Hierarchical | 0.152 | 0.444 | 0.254 | 0.429 | 12.7% |
| GIT-base-Direct Vision to Language | 0.006 | 0.174 | 0.002 | 0.153 | NA |
| ViT-T5small-Direct Vision to Language | 0.220 | 0.429 | 0.289 | 0.419 | NA |

We will ignore the direct VLM GIT base model in our discussions. The model is producing incoherent gibberish text, see Figure 9. This model is clearly under trained for the specific task. The problem does not lie with training time, as the model was early stopped due to overfitting, see Figure 10. Lowering the learning rate and other approaches like learning rate scheduler, weight decay etc. didn't work. I hypothesise that pre-training on text corpus is insufficient.

Figure 9 presents two examples, the first one with grey hooded sweatshirt is an example where both the direct VLM model and the MTL-Hierarchical model are successful. The MTL-Hierarchical model was able to successfully identify two "true" facts that were not present in ground truth – the jacket has sporty look and it has 2 pockets. Although the model does hallucinate about zip closure. While, the VLM model also identifies the new "true" fact about number of pockets, it also hallucinates about the material and misclassifies it as cotton.

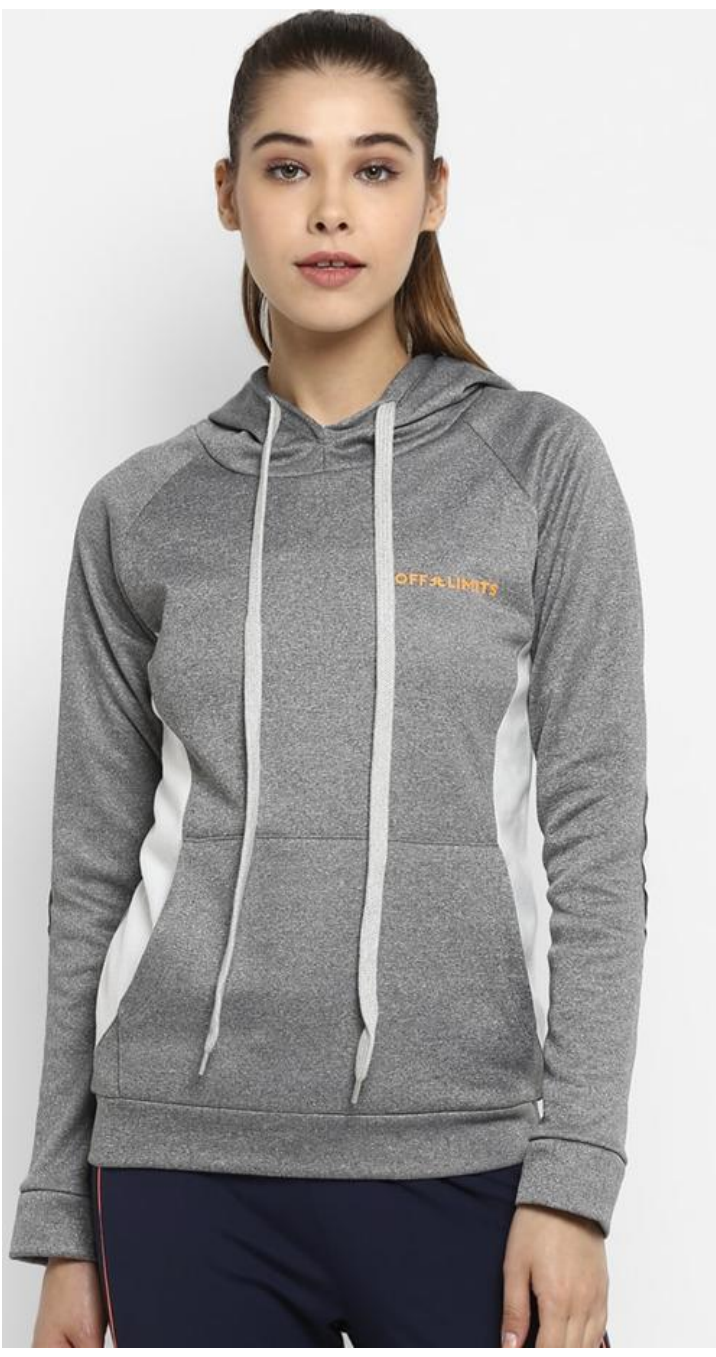

**Ground Truth**
**Title:** Women Grey Melange Solid Hooded Sweatshirt
**Description:** Grey Melange solid sweatshirt, has a hood, long sleeves, straight hem. The model (height 5'8") is wearing a size S. Material: Polyester Machine Wash",

**Multi Task Learning - Hierarchical**
**Title:** Women Grey Solid Sporty Jacket
**Description:** Grey solid sporty jacket, has a hooded, 2 pockets, zip closure, long sleeves, straight, polyester lining. The model (height 5'8") is wearing size S. Material: Polyester Machine Wash",

**Direct Vision-to-Language Model – ViT-T5small**
**Title:** Women Grey Solid Hooded Sweatshirt
**Description:** Grey solid sweatshirt has a hooded, 2 pockets, long sleeves, straight hem. The model (height 5'8") is wearing a size S. Material: Cotton

**Direct Vision-to-Language Model – GIT base**
**Title:** greyange sport jacket
**Description:** grey sweatshirt has hooded na, pockets zip, closure long, sleeves zip, hem model height ' ) wearing sizeater :ester wash wash similar",

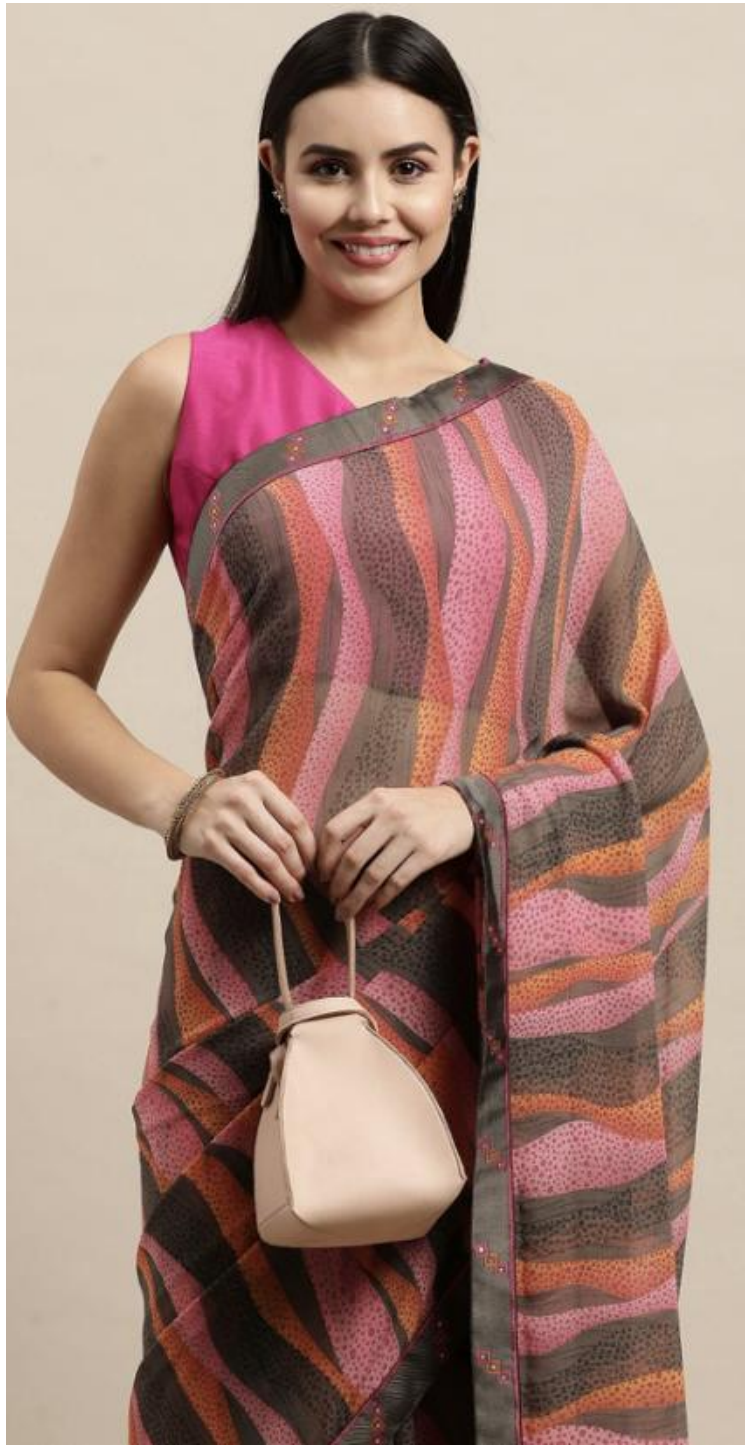

**Ground Truth**
**Title:** Pink & Grey Abstract Print Saree
**Description:** Design Details Pink and grey saree Abstract printed saree with woven design border. The saree comes with an unstitched blouse piece. The blouse worn by the model might be for modelling purpose only. Check the image of the blouse piece to understand how the actual blouse piece looks like. Georgette, Machine-wash. Length: 5.5 metres plus 0.8 metre blouse piece Width: 1.06 metres (approx.)",

**Multi Task Learning - Hierarchical**
**Title:** Pink, Geometric, Printed Pure Cotton Saree
**Description:** Design Details Pink and pink saree. Solid, geometric printed, printed printSilk blendDry CleanLength: 5.5 metres plus 0.8 metre printBlouse: 1.06 metres (approx.)",

**Direct Vision-to-Language Model – ViT-T5small**
**Title:** Pink Floral Printed Saree
**Description:** Design Details Pink printed saree with printed border. Has zari detail. The saree comes with an unstitched blouse piece. The blouse worn by the model might be for modelling purpose only. Check the image of the blouse piece to understand how the actual blouse piece looks like. Length: 5.5 metres plus 0.8 metre blouse piece Width: 1.06 metres (approx.)

**Direct Vision-to-Language Model – GIT base**
**Title:** & ethnic printed george sar
**Description:** and sar ethnicee printedee printedee printedee printedee printedee printedee printedee printedee printedee printedee printedee solidee solidee border sar comes ansti blousethe worn the might for purpose. the of blouse by model be modelling only check image the piece understand the piece understand the piece understand the piece understand the piece understand the piece understand the piece understand the piece understand the piece understand the piece understand the of blouse to how actual piece like length 5 5 5 5 5 5 5 plus. metre piece :. metres 0 8 blouse width 1 06",

*Figure 9: Qualitative comparison of text outputs of different models on test set. The blue highlighted text are the facts present in ground truth. The green highlighted facts are not present in ground truth but are actually true new facts found by models. The red highlighted facts are hallucinations*

The second example of pink and grey saree is an interesting case. This is exactly what we trying to prove in our analysis. The MTL-Hierarchical model is factually true but it lacks the fluency. This happens as we have encountered a rare case where most of the attributes were unknown. This confused the text generator. The text generator is clearly trying to “emphasise” the known parameter of pink color and pattern type of print. This is problem for smaller language model like T5-Small, causing it to prioritize inserting the factual keywords over generating elegant prose.

```
"mtl_attributes": {
  "Neck": "Unknown",
  "Sleeve Length": "Unknown",
  "Print or Pattern Type": "Geometric",
  "Type": "Unknown",
  "Hemline": "Unknown",
  "Pattern": "Printed",
  "Length": "Unknown",
  "Sleeve Styling": "Unknown",
  "Ornamentation": "Unknown",
  "Occasion": "Daily",
  "Fabric": "Unknown",
  "Fit": "Unknown",
  "colour": "Pink"
},
```

However, the VLM model fails silently. The phrasing is fluent but the details are all but true. Nor the print is of “Floral” Design, neither does the saree has printed border or zari detailing. This is very disastrous failure for real world applications, as a reviewer may miss the correction due to fluent phrasing. The incoherent text generated by the model is actually a better problem to have. It’s eye catching and lowers the chance of omission during manual review.

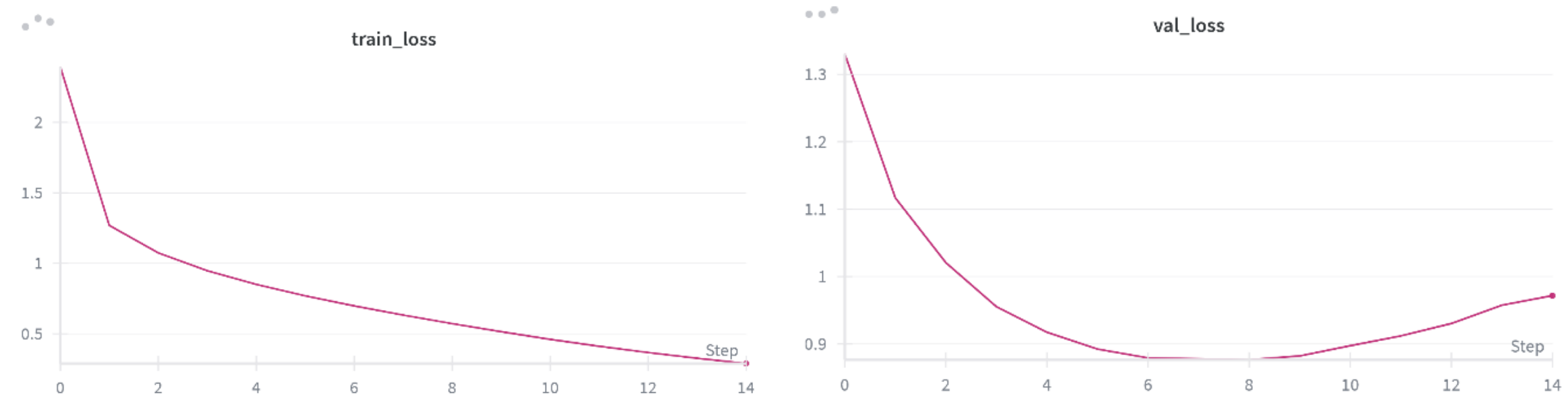

*Figure 10: Training and Validation Set Loss for direct Vision-to-Language model – GIT base. The best model saved at epoch 7 is used for analysis. Graph demonstrates the model has sufficient training time and overfitting in later epochs.*

## 6.5 Ablation Study – Joint Loss Function Weights

To validate the architectural claims, the weights of individual loss components – price, attributes & text generation losses- were varied. The weights were varied considering three scenarios – Balanced, Task Focused and Knock-Out. The results presented in Table 7 verifies the proposed claims. The data presented in Table 7 consists of evaluation run on 1422 test set samples – never seen in training. The table only shows Hallucinate Rate, Attribute F1 score and price MAE; for more metrics refer to Appendix B.

```
WEIGHT_COMBINATIONS = [
    # --- 1. Balanced Scenarios ---
    {'price': 0.3, 'attributes': 0.3, 'text': 0.4}, # Perfectly even baseline
    {'price': 0.4, 'attributes': 0.4, 'text': 0.2},  # Balanced, prediction-focused

    # --- 2. Task-Focused Scenarios ---
    {'price': 0.6, 'attributes': 0.2, 'text': 0.2},  # Price-focused
    {'price': 0.2, 'attributes': 0.6, 'text': 0.2},  # Attribute-focused
    {'price': 0.1, 'attributes': 0.1, 'text': 0.8},  # Text-focused
    {'price': 0.1, 'attributes': 0.4, 'text': 0.5},  # Attribute + Text focused

    # --- 3. "Knock-Out" Ablation Scenarios ---
    {'price': 0.0, 'attributes': 0.5, 'text': 0.5},  # No Price Task: Does price-awareness help other tasks?
    {'price': 0.5, 'attributes': 0.0, 'text': 0.5},  # No Attribute Task: Can the model still perform without fine-grained labels?
    {'price': 0.5, 'attributes': 0.5, 'text': 0.0},  # No Text Task: Can we learn a good HVR without a generation objective?
]
```

When the attribute weight is set to 0, and remaining is distributed equally to price and text loss, a hallucination of 67% is observed. This is significant discovery as the hallucinate rate is order of magnitudes higher than the cases. Also, as expected, the attribute F1 score is near zero – 0.017. This indicates the model has learned to predict the attributes correctly. This result reveals that attributes prediction acts a powerful regularizer for factual grounding and supports our hierarchical grounding proposal.

*Table 7: Results of ablation study for varying the weights of individual loss components*

| Sl. No. | Loss Function Coefficients | | | Price MAE | Price $R^2$ | Attribute F1 | Hallucination Rate (%) |
|---|---|---|---|---|---|---|---|
| | Price | Attribute | Text | | | | |
| **1** | **0** | **0.5** | **0.5** | **2998.2** | **-1.25** | **0.302** | **6.7** |
| 2 | 0.1 | 0.1 | 0.8 | 1011.6 | 0.50 | 0.272 | 7.5 |
| 3 | 0.1 | 0.4 | 0.5 | 1075.1 | 0.48 | 0.285 | 6.6 |
| 4 | 0.2 | 0.6 | 0.2 | 1083.6 | 0.45 | 0.276 | 5.9 |
| 5 | 0.3 | 0.3 | 0.4 | 1008.1 | 0.53 | 0.264 | 5.8 |
| 6 | 0.4 | 0.4 | 0.2 | 1043.8 | 0.46 | 0.267 | 6.4 |
| **7** | **0.5** | **0** | **0.5** | **1076.0** | **0.44** | **0.017** | **67.0** |
| 8 | 0.5 | 0.5 | 0 | 1023.0 | 0.45 | 0.264 | NA |
| 9 | 0.6 | 0.2 | 0.2 | 1036.2 | 0.47 | 0.223 | 6.7 |

Also, when the price weight is set to 0, and remaining is distributed equally to attribute and text loss, a Price $R^2$ of -1.25 is observed. This observation follows the expectation. The model is predicting a meaningless price value. However, the model achieved the highest attribute F1 score. It means that visual feature to attribute mapping can be learned effectively without including the price task. However, we can see from Table 3 that multi-tasking approach outperformed the siloed attribute learning approach on validation set. Thus, although not strictly necessary, the coupling of price prediction enables the model to learn subtle, non-obvious visual cues that are not listed in the attribute list like quality and style.

The case where text generation loss is set to zero shows that the removing the text loss has no significant negative impact on price or attribute prediction. However, the result from case 6 suggests that adding a little weight to text loss does slightly improves both price $R^2$ value and attribute F1 score.

For overall performance, case of {'price': 0.3, 'attributes': 0.3, 'text': 0.4} is selected as the final model. This case achieves lowest price MAE and lowest hallucination rate with a decent attribute F1 score of 0.264.

# 7. Conclusion

The automated e-commerce listing needs to address two challenges. First, the fragmented learning associated with siloed models trained to perform independent tasks. Second, the factual inconsistency associated with direct vision-to-language models. This work recommends a multimodal, multitask, hierarchical architecture that tackles both these issues together. The multitask learning approach has demonstrated its superiority over the siloed models in both price prediction and attribute classification tasks. The proposed approach achieved 3.6% better R2 value in price prediction than the siloed model. Also, the model provided a 6.6% attribute F1 score improvement over siloed model. This improvement was possible as the model was able to learn subtle visual cues, such as craftmanship, that were not explicitly listed under attributes. The hierarchical approach of grounding the text decoder using predicted attributes slashed down the hallucination rate from 12.7% to 7.1%, an improvement of 44.5%. The system prompt embedded with model's own predicted attributes acts an strong regularizer for the text generator. However, this improvement in hallucination rate comes as a trade off for fluency. The proposed model architecture showed a 3.5% lower ROUGE-L score than the direct vision-to-language models. But, in commercial setting the factual accuracy much more important than the fluency. While, the direct vision-to-language model generated a fluent lie, the proposed architecture generated much shorter length and somewhat gibberish text when it was not sure of output. The hierarchical approach also reduces the latency by a factor 3.5 by constraining the autoregressive algorithm. In conclusion, the work recommends and verifies a powerful "predict then generate" principle for image description task. The applicability of this principle can be beyond the e-commerce domain such as generating medical reports from X-rays.

# 8. Limitations and Future Work

While our work successfully demonstrates the architectural benefits of a multi-task, hierarchical approach for generating e-commerce listings, we acknowledge several limitations that define important avenues for future research. These can be broadly categorized into the domains of task complexity, model architecture, and real-world evaluation.

## 8.1 The Inherent Limits of Visual-Only Price Prediction

Our most significant limitation is the inherent difficulty of the visual-only price prediction task. Our results, while showing a clear relative improvement for our architecture, also reveal a modest absolute $R^2$ score. This is largely due to the brand confounder: the primary driver of price in fashion is brand identity, a non-visual feature that our model, by design, cannot access. A plain t-shirt from a luxury designer and a fast-fashion retailer may be visually indistinguishable, yet their prices can differ by orders of magnitude. Our model's performance is therefore capped by this "information ceiling."

- Future Work (Multimodal Feature Fusion): A powerful next step is to move beyond a purely visual system. The rich visual representation learned by our HVR could be fused with other data modalities. For instance, concatenating our learned image embedding with categorical embeddings for brand or seller_id before feeding them into a price prediction head could dramatically improve accuracy, leading to a more commercially viable system.

## 8.2 Constraints of the Input Modality and Dataset

Our current model operates on a single "hero" image from a specific domain, which limits its understanding and generalizability.

- Single-View Understanding: The model only sees one perspective of a product. It cannot discern features on the back, see close-up texture details, or understand how a garment fits on a human model.
- Domain and Cultural Specificity: The model is trained exclusively on apparel data, from a specific geographic and cultural market – women clothing in India. It would not generalize to other product categories like electronics, and its understanding of stylistic attributes (e.g., Occasion='Formal') may be biased towards Indian fashion norms.
- Future Work (Multi-View and Multi-Domain Models): A clear avenue for future work is to extend the vision encoder to handle multiple input images (e.g., front, back, detail shots) or even short product videos. This would provide a more complete product understanding. Furthermore, exploring domain adaptation techniques to fine-tune the model on other e-commerce verticals, such as home goods or electronics, would be a valuable test of its generalizability.

## 8.3 Architectural and Knowledge Boundaries

Our model, while effective, is static and self-contained. It has no access to the dynamic, external knowledge that informs real-world commerce.

- Lack of Real-Time Knowledge: The model is unaware of current fashion trends, competitor pricing, stock levels, or customer sentiment. Its generated descriptions, while factually grounded in the image, may not be commercially optimized for the current market.
- Generative Model Scale: We used T5-Small for efficiency. While capable, it may not possess the fluency or world knowledge of today's state-of-the-art Large Language Models (LLMs), limiting the creativity and sophistication of the generated text.
- Future Work (Retrieval-Augmented Generation - RAG): A transformative next step would be to integrate a RAG pipeline. Before generating a description, the model could retrieve relevant, real-time context, such as snippets from top-rated customer reviews for similar products, articles on current fashion trends, or competitor product descriptions. This would ground the model's output not just in the image, but in the live, dynamic context of the market.
- Future Work (Scaling the Language Module): Replacing the T5-Small decoder with a much larger, more powerful LLM could significantly enhance the quality, fluency, and persuasive power of the generated text, unlocking a new level of performance.

## 8.4 Evaluation Horizons

Finally, our evaluation, while robust, is based on offline metrics.

- Internal vs. External Consistency: Our hallucination metric cleverly measures the model's *internal* consistency (text vs. predicted attributes). It does not, however, guarantee *external* consistency (text vs. the ground-truth image). A model could be perfectly consistent with its own incorrect predictions.
- Future Work (Human Evaluation and Live A/B Testing): To truly validate the model's utility, two further evaluation stages are necessary. First, a large-scale human evaluation study where annotators rate the factual accuracy of the generated descriptions against the product images. Second, and most decisively, a live A/B test on an e-commerce platform to measure whether listings generated by Opti-List lead to a statistically significant improvement in real-world business metrics, such as click-through rate, add-to-cart rate, and ultimately, conversion.

---

## Acknowledgments

The authors acknowledge the use of a large language model (LLM) as a writing and editing assistant in the preparation of this manuscript. The tool was specifically utilized for assistance with drafting, refining the language, and structuring the narrative of the **Introduction**, **Related Work**, and **Limitations and Future Work** sections. The core scientific contributions of this paper, including the **Methodology**, **Experiments**, **Results and Discussion**, and **Conclusion**, were conceived, executed, and written exclusively by the human authors. The authors assume full and final responsibility for all content in this paper, including the accuracy of the data, the validity of the claims, and the scientific conclusions drawn.

# Appendix A: Implementation Details

*Table A1: List of choices used for the study.*

| Parameter Category | Parameter | Value / Choice |
|---|---|---|
| **Hardware** | GPU | Single NVIDIA GeForce RTX 4050 (6 GB VRAM) |
| **Proposed MTL Model** | **General** | |
| | Vision Encoder | ViT-Base/16 (google/vit-base-patch16-224-in21k) |
| | Text Decoder | T5-Small (t5-small) |
| | Optimizer | Adam |
| | Learning Rate | 1e-4 |
| | Batch Size | 16 |
| | Epochs | 10 |
| | **Loss Weights** | |
| | $\alpha$ (Attributes) | 0.4 |
| | $\beta$ (Price) | 0.1 |
| | $\gamma$ (Text) | 0.5 |
| **Baselines** | **Siloed Models** | |
| | Architecture | ViT-Base/16 (Same as proposed, trained individually) |
| | Batch Size | 16 |
| | Epochs | 10 |
| | Learning Rate | 1e-4 |
| | **Direct VLM (ViT-T5)** | |
| | Architecture | ViT-Base/16 + T5-Small |
| | Batch Size | 16 |
| | Epochs | 10 |
| | Learning Rate | 1e-4 |
| | **Direct VLM (GIT)** | |
| | Architecture | GIT-Base (microsoft/git-base) |
| | Batch Size | 12 |
| | Epochs | 15 |
| | Learning Rate | 2e-5 |
| **Inference Configurations** | Search Strategy | Beam Search |
| | Number of Beams | 4 |
| | Max Generation Length | 128 Tokens |
| | Input Prompt Length | Max 64 Tokens |

# Appendix B: Ablation Study Details

Here, we present the comprehensive results of the ablation study on the loss function weights, evaluated on the 1422-sample test set. This table expands upon the summary in Section 6.5 to include text fluency and price regression error metrics, demonstrating the interplay between task priorities.

*Table B1: Comprehensive results of the loss weight ablation study.*

| Sl. No. | Loss Function Coefficients | | | Price MAE | Price RMSE | Price R2 | Attribute F1 | BLEU | ROUGE-1 | ROUGE-2 | ROUGE-L | Hallucination Rate (%) |
|---|---|---|---|---|---|---|---|---|---|---|---|---|
| | Price | Attribute | Text | | | | | | | | | |
| 1 | 0 | 0.5 | 0.5 | 2998.2 | 4019.1 | -1.25 | 0.302 | 0.1568 | 0.4216 | 0.2418 | 0.404 | 6.7 |
| 2 | 0.1 | 0.1 | 0.8 | 1011.6 | 1884.2 | 0.50 | 0.272 | 0.1572 | 0.4243 | 0.2418 | 0.4058 | 7.5 |
| 3 | 0.1 | 0.4 | 0.5 | 1075.1 | 1930.9 | 0.48 | 0.285 | 0.1511 | 0.4105 | 0.2371 | 0.3953 | 6.6 |
| 4 | 0.2 | 0.6 | 0.2 | 1083.6 | 1988.5 | 0.45 | 0.276 | 0.1469 | 0.3922 | 0.2274 | 0.3774 | 5.9 |
| 5 | 0.3 | 0.3 | 0.4 | 1008.1 | 1833.8 | 0.53 | 0.264 | 0.141 | 0.3911 | 0.2221 | 0.3745 | 5.8 |
| 6 | 0.4 | 0.4 | 0.2 | 1043.8 | 1975.8 | 0.46 | 0.267 | 0.1549 | 0.4092 | 0.2361 | 0.3939 | 6.4 |
| 7 | 0.5 | 0 | 0.5 | 1076.0 | 1998.2 | 0.44 | 0.017 | 0.1129 | 0.335 | 0.1777 | 0.3229 | 67.0 |
| 8 | 0.5 | 0.5 | 0 | 1023.0 | 1981.2 | 0.45 | 0.264 | 0.0011 | 0.0374 | 0.0011 | 0.0313 | NA |
| 9 | 0.6 | 0.2 | 0.2 | 1036.2 | 1954.4 | 0.47 | 0.223 | 0.1416 | 0.3996 | 0.2229 | 0.3838 | 6.7 |

---

# Appendix C: Per-Attribute Performance Analysis

To better understand the challenges of the attribute prediction task, we report the Macro F1-Score for each of the 12 selected attributes on the test set. As hypothesized in the main paper, inferred or texture-based attributes like 'Fabric' prove more difficult to classify than visually distinct attributes like 'Sleeve Length' and 'Neck' style.

*Table C1: Per-attribute Macro F1-Scores for the proposed multitask model.*

| Attribute | Macro F1 Score |
|---|---|
| Sleeve Length | 0.830 |
| Length | 0.569 |
| Pattern | 0.447 |
| Occasion | 0.441 |
| Hemline | 0.393 |
| Colour | 0.278 |
| Neck | 0.242 |
| Type | 0.215 |
| Print or Pattern Type | 0.214 |
| Fit | 0.201 |
| Ornamentation | 0.195 |
| Fabric | 0.185 |
| Sleeve Styling | 0.165 |

***Average Macro F1-Score: 0.3365***

# Appendix D: Additional Qualitative Examples

This section provides additional qualitative examples from the test set to supplement Figure 9.

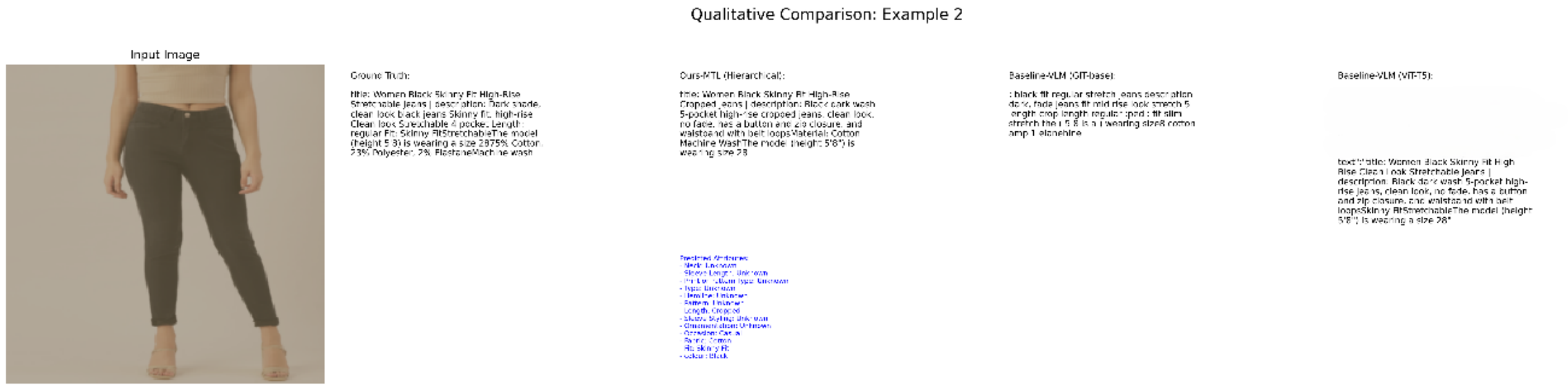

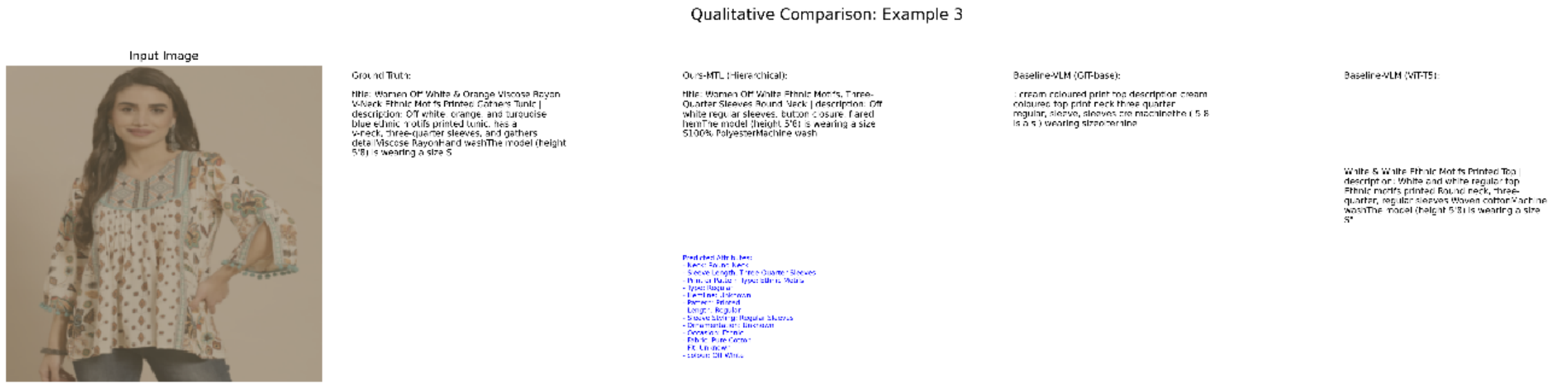

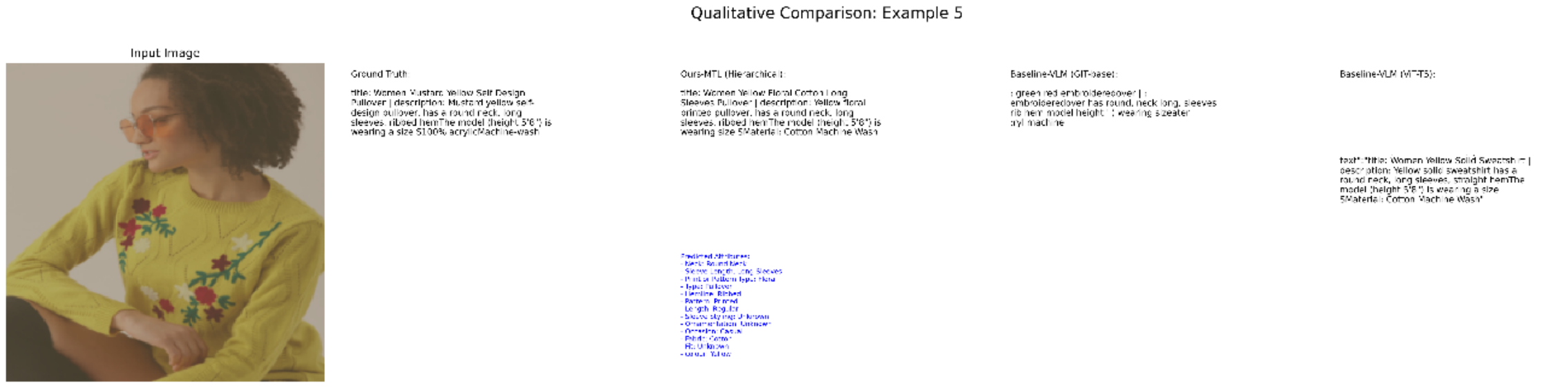

# Appendix E: Dataset Details and Preprocessing

This section details the origin, characteristics, and preprocessing pipeline of the dataset used for training and evaluating our models.

### E.1 Data Collection and Source

The dataset was curated from publicly available product listings on Myntra.com, a major Indian e-commerce platform specializing in fashion and apparel. Data was collected in early 2024 using an automated script. The collection process was rate-limited and scheduled during off-peak hours to ensure minimal impact on the source's infrastructure. Only publicly accessible information was gathered for each product. The raw dataset contained 14,214 unique listings for women's apparel.

### E.2 Initial Data Characteristics

The raw dataset included the following primary fields:

- p_id: A unique product identifier.
- name: The product title.
- price: The price of the product in Indian Rupees (INR).
- colour, brand: Basic product attributes.

- img: The URL to the primary product image.
- ratingCount, avg_rating: User rating information.
- description: The textual product description.
- p_attributes: A string-formatted dictionary containing a rich but unstructured set of over 50 product attributes.

Initial analysis revealed that the ‘ratingCount’ and ‘avg_rating’ fields had significant missing values (7,684 missing rows, ~54% of the dataset), while all other core fields were complete.

**E.3 Preprocessing Pipeline**

To prepare the data for the multitask model, the following preprocessing pipeline was executed:

1. Attribute Parsing: The p_attributes column, which stored attributes as a string, was parsed into a structured dictionary format for each product using Python's ast.literal_eval. This step made the rich attribute data machine-readable.
2. Attribute Selection and Extraction: Based on an analysis of the most frequent and visually identifiable attributes across the dataset, a final set of 12 key attributes was selected for the model's prediction tasks. These attributes were extracted from the parsed dictionaries into their own dedicated columns in the dataframe. The selected attributes are listed in Table 4.1 in the main paper.
3. Data Cleaning and Normalization:
   - For the selected attributes, string values of 'NA' were standardized to a consistent 'Unknown' category to handle missing data explicitly.
   - The description and name fields were cleaned to remove HTML tags (e.g., <br>) and lowercased for consistency.
   - The p_id was used to create a local file path for each corresponding image, decoupling the dataset from online URLs.
4. Feature and Target Finalization:
   - Irrelevant columns for the primary task, such as the original p_attributes string, the parsed attributes_dict, and the product brand, were removed to create the final, clean dataset. The brand was explicitly removed to ensure the model learned price from visual cues rather than brand identity, as discussed in Section 8.1.
5. Data Splits: After cleaning, the dataset was partitioned into training, validation, and test sets using an 80-10-10 ratio, resulting in:
   - Training Set: ~11,371 samples
   - Validation Set: ~1,421 samples
   - Test Set: ~1,421 samples

# Appendix F: Computational Environment

All experiments were conducted in the following computational environment. Key library versions are listed to ensure precise reproducibility.

*Table F1: Key software libraries and versions used in the experiments.*

| Category | Library/Component | Version |
|---|---|---|
| **Core Frameworks** | PyTorch | 2.8.0 (+cu126) |
| | Transformers | 4.57.0 |
| | Scikit-learn | 1.7.2 |
| | XGBoost | 3.0.5 |
| **Data & Computation** | Python | 3.10+ (Recommended) |
| | NumPy | 2.3.3 |
| | Pandas | 2.3.2 |
| | Pillow (PIL) | 11.3.0 |
| **Evaluation** | Evaluate | 0.4.6 |
| | Rouge Score | 0.1.2 |
| | NLTK | 3.9.1 |
| **Experimentation** | Weights & Biases (wandb) | 0.22.0 |
| | PyYAML | 6.0.2 |
| | TQDM | 4.67.1 |